# Dynamic Functional Connectivity and Graph Convolution Network for Alzheimer's Disease Classification

Xingwei An
Academy of Medical Engineering and Translational Medicine, Tianjin University, Tianjin, China 300072
anxingwei@tju.edu.cn

Yutao Zhou
Academy of Medical Engineering and Translational Medicine, Tianjin University, Tianjin, China 300072
harlanzhou@tju.edu.cn

Yang Di
Academy of Medical Engineering and Translational Medicine, Tianjin University, Tianjin, China 300072
dy_aquarius@tju.edu.cn

Dong Ming
Academy of Medical Engineering and Translational Medicine, Tianjin University, Tianjin, China 300072
richardming@tju.edu.cn

## ABSTRACT
Alzheimer's disease (AD) is the most prevalent form of dementia. Traditional methods cannot achieve efficient and accurate diagnosis of AD. In this paper, we introduce a novel method based on dynamic functional connectivity (dFC) that can effectively capture changes in the brain. We compare and combine four different types of features including amplitude of low-frequency fluctuation (ALFF), regional homogeneity (ReHo), dFC and the adjacency matrix of different brain structures between subjects. We use graph convolution network (GCN) which consider the similarity of brain structure between patients to solve the classification problem of non-Euclidean domains. The proposed method's accuracy and the area under the receiver operating characteristic curve achieved 91.3% and 98.4%. This result demonstrated that our proposed method can be used for detecting AD.

## CCS Concepts

• **Applied computing**➞**Life and medical sciences**➞**Computational biology**➞**Imaging**

## Keywords
Alzheimer's disease (AD); graph convolutional network (GCN); functional connectivity (FC); classification

## 1. INTRODUCTION
Alzheimer's disease (AD) is an irreversible, progressive neurodegenerative disease that primarily affects the elderly and is the most prevalent form of dementia [1]. The initial clinical manifestations of the patients are a variety of common symptom such as memory loss, speech impairment and cognitive deficit, which severely affected and limited personal daily life and even pose a grave threat of patient with the progression of the disease. It is estimated that with the increasing aging of the global population, 1 out of 85 people will be affected by AD in the future [2]. Therefore, the diagnosis of AD is very crucial.

Now, the approach commonly used in this field is to combine resting-state functional magnetic resonance imaging (rs-fMRI) with machine learning and the stationary functional connectivity (sFC). Deep learning methods are also applied to identify and detect AD.

However, most of the previous studies believe that these methods have some drawbacks. First, machine learning fails to consider the association between patients. Second, the traditional deep neural network CNN cannot effectively solve the graph structure, since the brain is considered as a complex network with small-world attributes. Deep learning method contain a large number of parameters, which will give rise to long training time and high requirements for computer hardware. Third, sFC cannot reflect time-varying dynamic behavior and ignore the local dynamic changes of the brain during the whole time series [3].

Graph convolution network (GCN) model was proposed to handle the problem of non-Euclidean domains, which can take into account the similarity between different individuals by aggregating adjacency subjects [4, 5]. Compare with other deep learning methods, GCN has simple structure, less parameters and training time. It is conductive to assisting doctors to make decisions timely and accurately. Dynamic functional connectivity can effectively focus on changes in the relationships between brain regions in different time sub-segments.

In this paper, we propose a novel method to classify AD based GCN model and dynamic functional connectivity. Specifically, we combine with different type of features such as amplitude of low-frequency fluctuation (ALFF), regional homogeneity (ReHo) and thresholding dynamic functional connectivity (dFC) as feature set was utilized to conduct a classification study of AD with full consideration of individual similarity and data association between subject's brain information. Our method was proved that can significantly improve prediction performance and execution speed.

## 2. MATERIALS AND METHODS

### 2.1 Data Acquisition

The rs-fMRI time-series data used in this paper were collected from Xuanwu Hospital, Beijing, China. Patients were provided written informed consent. There are total of 423 scans which contained 120 AD and 303 normal controls (NC) from 317 subjects, including 106 AD and 211 NC subjects. Then, we choose 246 (98 AD, 148 NC) scans from 204 patients that contain 94 AD and 110 NC subjects. It's clearly that some subjects scanned twice or more times, separated by at least one year. 177 scans were excluded for different total time points.

### 2.2 Data Preprocessing

All the resting state functional images were preprocessed and analyzed respectively by Data Processing Assistant for Resting-state fMRI (DPARSF 4.2, http://www.restfmri.net/forum/DPARSF) and Statistical Parametric Mapping (SPM12, http://www.fil.ion.ucl.ac.uk/spm)[6]. The preprocessing operations are as follows:

(1) remove time points,

(2) slice timing correction,

(3) head motion realignment. 20 AD and 3 NC scans with a max head motion over 3.0mm translation or 3° rotation were discarded,

(4) normalize

(5) smooth

(6) detrend

(7) nuisance covariates regression

(8) temporal filter ranging from 0.01-0.08Hz.

Notably, regional homogeneity (ReHo) were obtained without spatial smoothing. The subjects' information of this study as shown in Table 1.

**Table 1. Subjects' information of this study**

| Group | AD | NC |
|---|---|---|
| gender(F/M) | 47/31 | 79/66 |
| age(mean±std) | 71.1±9.7 | 64.64±8.53 |

### 2.3 Feature Extraction and Selection

Choosing the appropriate feature set can improve the performance of the classifier and reduce the training time. This section will introduce feature extraction and selection from four aspects.

#### 2.3.1 ALFF and ReHo Feature Extraction

Two sample T-test was used to extract features subsets from the features sets of ALFF and ReHo, we choose false discovery rate (FDR) correction and set $p < 0.05$ and cluster $size > 50$. Finally, we remained 4 ALFF and 5 ReHo features.

#### 2.3.2 Dynamic FC Construction

In this section, brain was parceled into 116 ROIs by Automated Anatomical Labeling (AAL) atlas. Time series were extracted from each region. To capture temporal variability, the entire time course was split into multiple sub-segments by sliding window approach. We can get K sub-segments by this formula: $K = \left[\frac{T-L}{s}\right] + 1$, where $T$ means the length of time points, $L$ means the length of the sliding window, $s$ means the length of step and $K$ means the number of segments. Pearson's correlation coefficient (PCC) was used to construct functional connectivity matrix's element: $c_{ij}^l = corr\{x_i^l, x_j^l\}$ between region $x_i^l$ and region $x_j^l$ in $l$-th sliding window. Then we use Fisher's $z$ transformation to normalized $r$ value and obtain a 116*116 symmetric matrix $C$.

#### 2.3.3 Dynamic FC Features Extraction

In current study, our original FC feature set include a large number of features. Thresholding operation was performed to simplify feature selection stage. We assign the same specific threshold $\tau$ for each subject and obtain a new matrix $M_a^{(l)} = \left[m_{ij}^{(l)}\right]_{n \times n}$ by

$$m_{ij}^{(l)} = \begin{cases} \left|c_{ij}^{(l)}\right|, & \left|c_{ij}^{(l)}\right| > \tau \\ 0, & else \end{cases} \quad (1)$$

where $m_{ij}^l$ denotes sub-segment the $l$-th connection strength between ROI $i$-th and ROI $j$-th of the subject $a$-th. Next we change $M_a^{(l)}$ into $A_a^{(l)}$ by binary operation for creating similarity adjacency matrix. Therewith we remove matrix $M$'s lower triangles and the main diagonal values and then retain the upper triangles 6670 values to collapse these values into a vector. The advantage of dynamic FC can sufficiently capture localized properties in different time series. The same feature was accumulated in different segments as follows:

$$m'_{ij} = \sum_{l=1}^{K} m_{ij}^l, \; m'_{ij} \in M_a \quad (2)$$

**Figure 1. The pipeline of GCN model for AD classification**

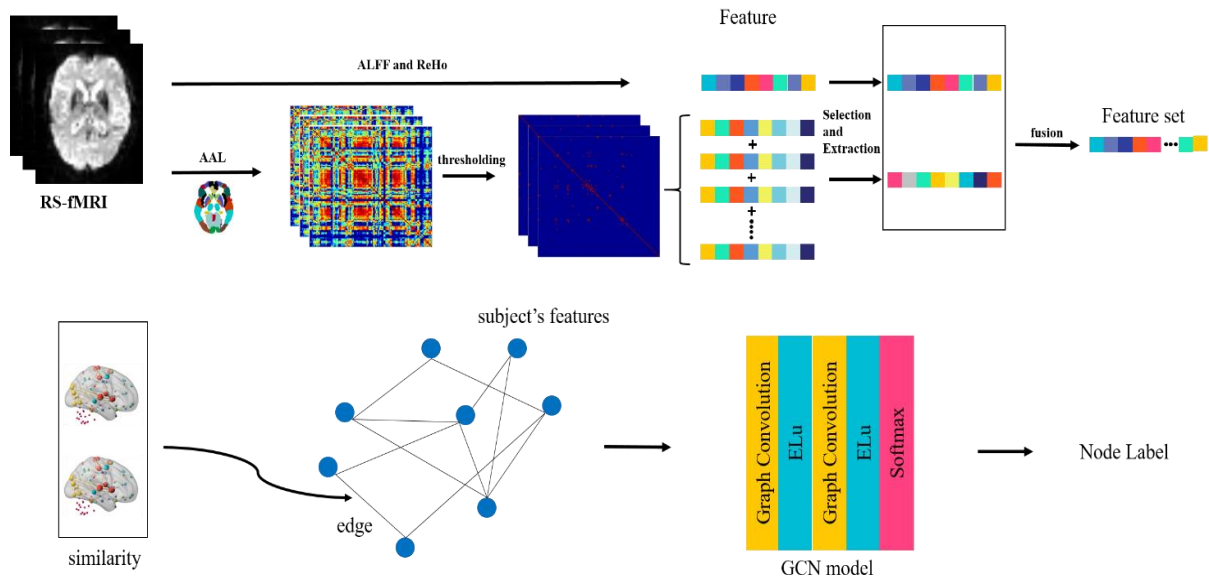

*2.3.4 Dynamic FC Features Selection*

After the previous operation, two-step approach which significantly improved speed and efficiency was employed for feature selection. First, we take random forest and obtain the most importance 10 features. Second, we use a recursive feature elimination SVM (RFE-SVM) which is backward elimination method that starts with a full set of all features and then remove the most irrelevant features one by one. Final, 9 features were remained for training classifier. The feature subset of ALFF、ReHo and dFC were set as a feature matrix $X \in \mathbb{R}^{n \times m}$ each row represents a subject's $m$ feature vectors.

## 2.4 Graph Convolution Network

GCN is used to solve non-Euclidean problems, such as event graphs, knowledge graphs, brain network and so on. We define an undirected graph $G(V, E, S)$ with $N$ nodes to describe personal network classification that each subject is represented by a node $v_i \in V$, $e_{ij}$ denotes edge weight which the similarity of between individual matrix $M_i$ and individual matrix $M_j$. In order to construct a similarity adjacency matrix $S$, $s_{ij} \in S$ between the subject $i$ and $j$ in terms of these formulas:

$$s_{ij} = 1 - \frac{\sum |A_i - A_j|^2}{\sum |A_i|^2} \quad (3)$$

$$S(s_{ij}) = \begin{cases} 1, s_{ij} < t \\ 0, s_{ij} \geq t \end{cases} \quad (4)$$

where $\Sigma$ denotes the sum of matrix's element and adjacency matrix between different ROI regions were utilized to compute similarity adjacency matrix S.

GCN take the neighbors of subject into account which can associate with other subject's information to make prediction. In this research, we finish classification for AD and NC as follows:

$$H^{(l+1)} = f(H^{(l)}, S) \quad (5)$$

$$\hat{S} = S + I \quad (6)$$

$$f(H^{(l)}, S) = \sigma(\hat{D}^{-\frac{1}{2}} \hat{S} \hat{D}^{-\frac{1}{2}} H^{(l)} W^{(l)}) \quad (7)$$

where $H^{(l)}$ means feature matrix that $H^{(0)} = X$, $S$ means similarity adjacency matrix, $I$ means identity matrix, $W^{(l)}$ means the weight matrix of $l$th layer, $\sigma$ is an activate function. The pipeline of GCN model for AD classification is illustrated in Figure 1.

## 3. RESULTS AND DISCUSSION

The model we used in this paper consist of two graph convolution layers. Each layer followed by an exponential linear unit (ELU). A softmax was used as output layer. The settings of the hyper-parameters during the training stage are as follow: learning rate is 0.06, dropout rate is 0.5, weight decay is 0.0005, the number of epoch is 150, the number of input layer's neurons is 18 and the number of hidden layer's neurons is 16. We adopt the Adam algorithm and the cross-entropy as optimizer and loss function, respectively. The sliding window length L is 39 time points with the step size s of 5. The classification accuracy (ACC), precision (PRE), recall (REC), F1 score (F1) and the area under the receiver operating characteristic curve(AUC) were used as evaluation criteria to evaluate the classification performance.

### 3.1 Performance of the Different Feature Sets

In this part, we verify from three different types of features and different numbers of features of the same type in order to obtain the optimal feature combination as input.

As shown in Table 2, different types of the feature combinations have different classification results. The 9 FC features achieved the improvement of 7%, compared with 6670 FC features. It is indicated that 9 FC features contained some discriminative features that played an important role for recognizing AD. In addition, the performance of combined features is generally better than the performance of single feature. More specifically, we proposed a method that concatenate three type of features to execute the classification task yielded the accuracy up to 91.3%, which outperform than other the combination of feature. Due to ALFF、ReHo and FC are heterogeneous, combining them can provide complementary information each other. Finally, we found ALFF and ReHo features including fusiform gyrus, right insula, right dorsolateral superior frontal gyrus, right inferior temporal gyrus, left medial orbitofrontal gyrus and cerebellum, which manifest that the cerebellum might be related to cognition [7]. Results proved that the cerebellum can provide crucial information for the classification of AD.

Table 2. Classification performance of different type and number of feature combinations

| Method | | ACC (%) | PRE (%) | REC (%) | F1 (%) | AUC (%) |
|---|---|---|---|---|---|---|
| type | number | | | | | |
| ALFF | 4 | 82.6 | 77.8 | 77.8 | 77.8 | 84.9 |
| ReHo | 5 | 78.2 | 75.0 | 66.7 | 70.6 | 85.7 |
| FC | 9 | 87.0 | 80.0 | **88.9** | 84.2 | 97.6 |
| FC | 6670 | 80.0 | 64.7 | 78.6 | 71.0 | 86.4 |
| ALFF+ReHo | 9 | 82.6 | 77.8 | 77.8 | 77.8 | 85 |
| ALFF+FC | 13 | 86.9 | 80.0 | **88.9** | 84.2 | 96.8 |
| ReHo+FC | 14 | 78.3 | 75.0 | 66.7 | 70.6 | 85.7 |
| **ReHo+ALFF+FC** | 18 | **91.3** | **100** | 77.8 | **87.5** | **98.4** |

## 3.2 Compare with State-of-the Art Method

Based on the previous description, in order to test the robustness of our method, we readjust the data distribution of the training set, validation set and test set from 6: 2: 2 to 3: 1: 6, which is different from traditional data set distribution ratio. We believe that this method may be suitable for unlabeled data, which can considerably reduce the cost of labeling data.

In current studies, researchers have taken many different ways to classify and predict AD and NC, they hope to get a trade-off between training speed and model performance. Wang et al. [8] introduced the spatial-temporal information into deep neural network model that model employed convolution component, recurrent component and long short-term memory (LSTM) to process the dependence between time and space. Their method's accuracy and AUC achieved 90.28% and 89.78%, respectively.

Song et al. [9] achieved an accuracy of 88.7% using the graph neural network to classify 40 late mild cognitive impairment (MCI) and 67 NC. This method utilized construct functional connectivity network with static, dynamic and high-level FC. The result of this model show that classifier had pretty good performance.

SVM is also widely used in research in this field. Bi et al. [10] proposed a clustering evolutionary random forest for feature extraction and selection. Their accuracy rate was 86.2%.

The proposed method in this study has some advantages. On the one hand, compared with traditional deep learning models, our model has a simple architecture and a short training time. On the other hand, compared with the traditional machine learning method, our method can consider the similarity between different subjects and classify and predict the subjects as a whole.

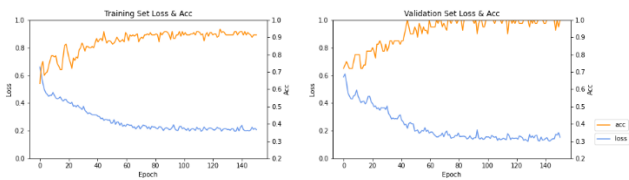

**Figure 2. The accuracy and loss of training and validation set**

## 4. CONCLUSION

In this paper, we propose a novel method based on GCN and dynamic connectivity for AD classification. In order to construct graph, three different types of features were fused to represent each vertices and edges were used to capture the similarity of brain structure between individuals. Result demonstrated that consider the structural similarity between different individual brains is helpful to improve the classification accuracy. This method can effectively promote training speed and model performance.

## 5. ACKNOWLEDGMENTS

The authors sincerely thank Xuanwu Hospital for the data provided to us, and all patients for their active cooperation. This work was supported in part by the National Key Research & Development Program of China (No.2017YFB1300302) and National Natural Science Foundation of China (No.81630051 and 61603269).